\documentclass[]{article}

\title{LightPanel: Active Mobile Platform for Dense 3D Modelling}
\author{Jonas Schuler, Reza Sabzevari and Davide Scaramuzza\\\\
	Robotics and Perception Group, University of Zurich, Switzerland}

\usepackage{amsmath,graphicx}
\usepackage{tabularx, booktabs, multirow}
\usepackage[breaklinks=true]{hyperref}

\def\mat#1{\mathchoice{\mbox{\boldmath$\displaystyle\tt#1$}}
{\mbox{\boldmath$\textstyle\tt#1$}}
{\mbox{\boldmath$\scriptstyle\tt#1$}}
{\mbox{\boldmath$\scriptscriptstyle\tt#1$}}}

\def\vec#1{\mathchoice{\mbox{\boldmath  $\displaystyle\bf#1$}}
{\mbox{\boldmath  $\textstyle\bf#1$}}
{\mbox{\boldmath  $\scriptstyle\bf#1$}}
{\mbox{\boldmath  $\scriptscriptstyle\bf#1$}}}

\newlength{\colwidth}
\begin{document}

\maketitle

\begin{abstract}
In this paper we introduce a novel platform for dense 3D modelling. 
This platform is an active image acquisition setup assisted with a set of light sources and a distance sensor.
The hardware setup is designed for being mounted on a mobile robot which is remotely driven to create accurate dense 3D models from out-of-reach objects.
For this reason, the object is actively illuminated by the imaging setup and \textit{Photometric Stereo} is used to recover the dense 3D model.
The proposed image acquisition setup, called \textit{LightPanel}, is described from design to calibration and discusses the practical challenges of using \textit{Photometric Stereo} under uncontrolled lighting conditions. 

\end{abstract}

\section{Introduction}
\label{sec:intro}
\begin{figure}[t]
	\centering
	\centerline{\includegraphics[width=10cm, trim = 20mm 50mm 20mm 50mm, clip]{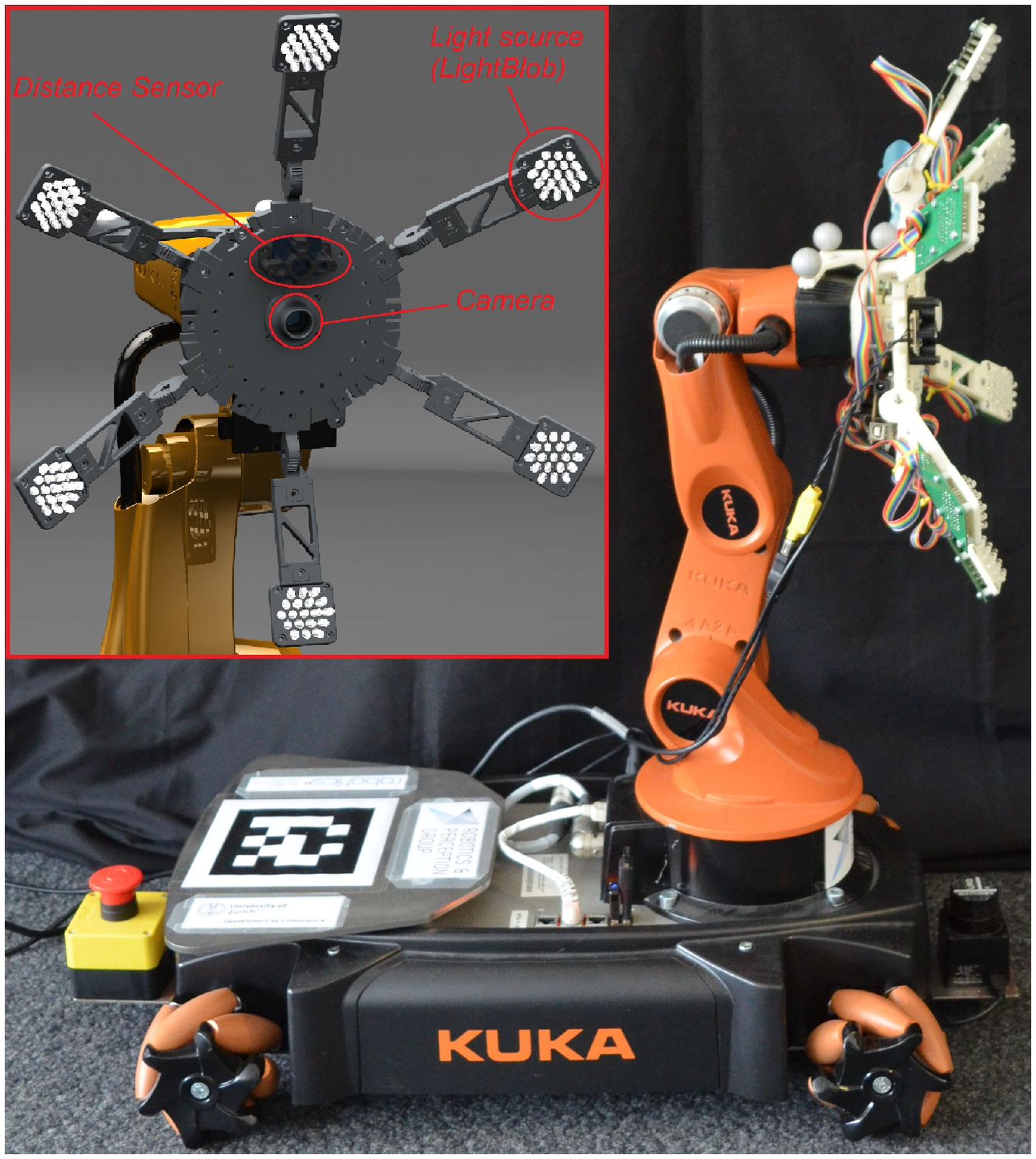}}
	\caption{The \textit{LightPanel} mounted on a KUKA youBot.}
	\label{fig:lightpanel_CAD}
\end{figure}

This paper presents a practical system for using \textit{Photometric Stereo} (PS) in robotics applications.
The hardware consists of a configurable image acquisition system, designed to be mounted on a mobile robot, i.e. KUKA youBot \cite{kukayoubot}.
The motivation of using such 3D reconstruction system is to accurately create 3D models of objects located in some places which are either dangerous or difficult for humans to reach in search and rescue scenarios.
Resulting dense 3D models can be used for critical tasks where the objects have to be manipulated remotely, i.e. inspection, part insertion, drilling, and so on.
The \textit{LightPanel} illuminates the object from different directions and takes images under varying lighting conditions from a fixed view point.
Such images are suitable for photometric dense reconstruction using PS.
\textit{LightPanel} can be easily tuned by changing the distance and angle of each light source with respect to the camera.
Reconfigurability of the \textit{LightPanel} allows illuminating objects in different distances for dense 3D reconstruction.
Fig.~\ref{fig:lightpanel_CAD} shows the \textit{LightPanel} and its components mounted on the KUKA youBot.

PS is a technique to estimate the surface points' orientations from multiple images under varying illumination, introduced by Woodham in~ \cite{woodham1980photometric}.
The Woodham's algorithm \cite{woodham1980photometric} has been enhanced in different directions.
Uncalibrated PS, for unknown lighting was introduced by Hayakawa \cite{hayakawa1994photometric} and improved in \cite{basri2007photometric, ackermann2012photometric, papadhimitri2013new}.
Further work to extend the algorithm with more complex lighting models for non-diffusive materials is highlighted in \cite{hernandez2008shadows, verbiest2008photometric, sun2007object}. 
Another way of improving the standard calibrated photometric stereo algorithm was presented in \cite{galo1996surface, jones2011head, tankus2005photometric} by adapting the formulas to a perspective camera model instead of the orthographic projection.
In \cite{galo1996surface, jones2011head, kozera2006noise} the assumption of a distant light source is relaxed.
The field of application is wide and reaches from industrial inspection \cite{landstrom2013sub} and quality control \cite{farooq2005dynamic} to archaeology with studies of ancient artefacts \cite{einarsson2004photometric}.
In \cite{willems2005easy}, a dome was introduced, where the object is placed in the centre and the light sources are fixed on the walls of the dome.
A frame for calibrating the illumination directions is placed around the object in \cite{einarsson2004photometric} with a hand-held light source.
In \cite{einarsson2004photometric} a big setup has to be installed in front of the object and in \cite{willems2005easy, debevec2000acquiring} the object has to be placed in a capturing tool.
Most similar image acquisition designs to our \textit{LightPanel} are introduced by Zhou \cite{zhou2010ring} and Malzbender \cite{malzbender2006surface} where the first real-time PS algorithm was implemented with a setup using wings equipped with LEDs.

Although PS reconstruction and its application is widely studied in the literature, the lighting condition is either controlled or a priori knowledge is considered in this regard. 
The \textit{LightPanel} hardware is low-cost and energy efficient which makes it suitable for using on a mobile robot.
This work investigates the practical challenges in employing PS reconstruction in environments with uncontrolled lighting condition for robotics applications.
In addition to introducing a mobile image acquisition system, this paper looks into two fundamental questions regarding the lighting condition:

\begin{itemize}
	\item How does the active illumination affect the reconstruction? How should the object be illuminated?
	\item Is the ambient light helpful for PS reconstruction or is not desirable at all?
\end{itemize}

In the following section, a brief introduction to photometric image formation and PS dense reconstruction is given. 
Afterwards, the proposed image acquisition hardware (i.e. \textit{LightPanel}) is introduced and supported by discussion on calibration the system.
Then, practical challenges to use \textit{LightPanel} are discussed and followed by the reconstruction results.

\section{Photometric Stereo}
\label{sec:ps}

\begin{figure}
	\centering
	\centerline{\includegraphics[width=0.9\linewidth, trim = 0mm 25mm 0mm 25mm, clip]{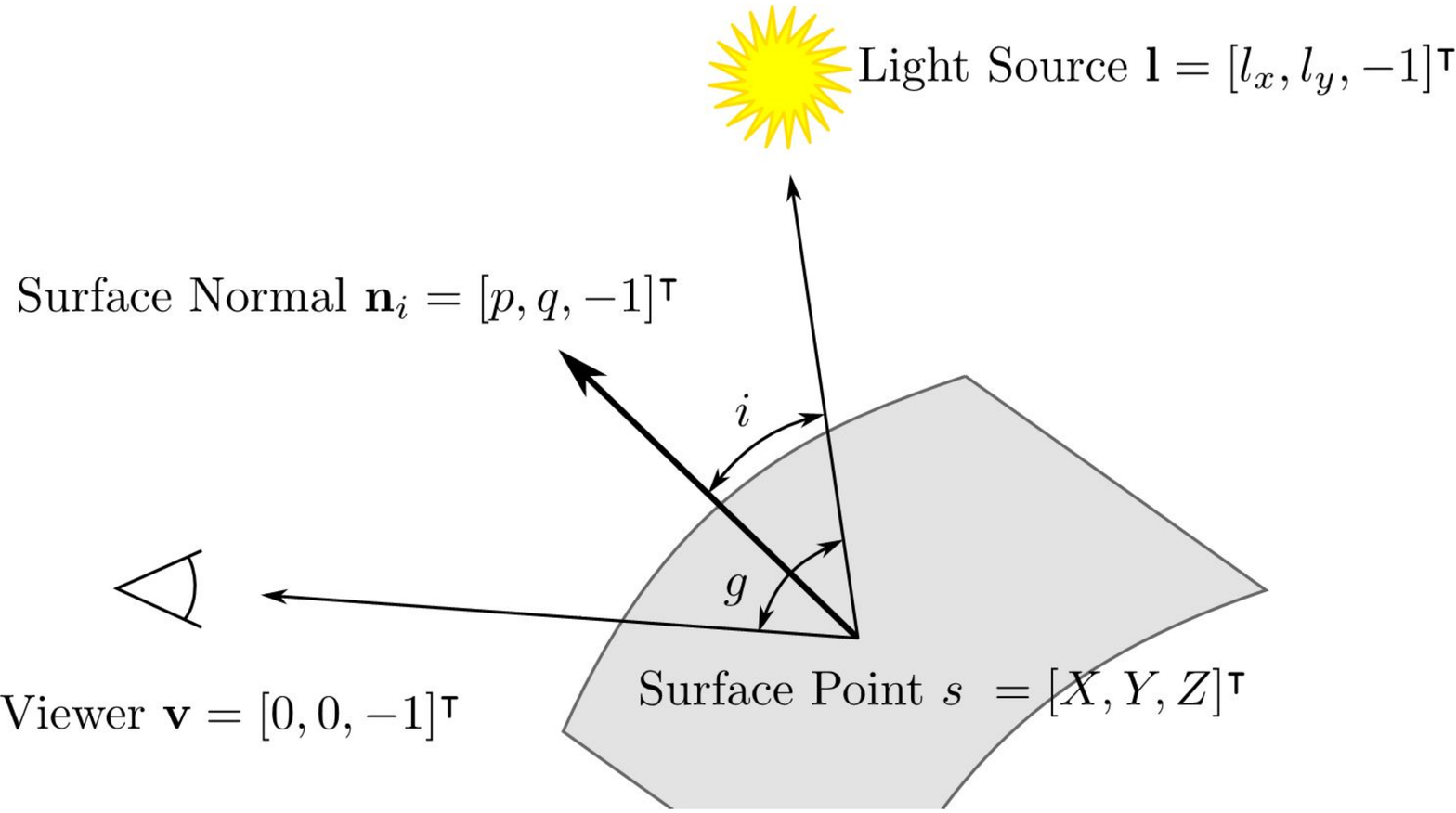}}
	\caption{Reflectance geometry: \textit{Incident} angle $i$ is between the light source ray and the surface normal and the \textit{phase} angle $g$ is between the light source ray and the viewer direction.}
	\label{fig:ref_geo}
\end{figure}

Standard PS methods assume that the camera and object are fixed and a single distant light source is moved during the image acquisition.
For this reason, the correspondence between the image points is given and can also be used for featureless objects \cite{angelopoulou2014evaluating}.
Since no features have to be matched, it is computationally cheaper than other reconstruction methods.
Assuming a perfect defuse surface, the observed intensity $\vec I(x, y)$ of a surface point $s$ is defined as:
\begin{equation}
\vec I(x,y) = \rho\cos(i) = \frac{\rho (1 + pl_x + ql_y)}{\sqrt{1 + p^2 + q^2}\sqrt{1 + l_x^2 + l_y^2}},
\label{eq:reflectance_function}
\end{equation}
where $i$ is the light incident angle and $\rho$ is the reflectance factor, also called albedo which is defined as diffuse reflectivity.
$p, q, l_x$ and $l_y$ are the components of vectors representing the surface normal and the light source direction, as depicted in Fig.~\ref{fig:ref_geo}.
In \eqref{eq:reflectance_function} the observed intensity does not depend on the viewer's direction.
Such relation is only valid for perfect diffuse surfaces, called Lambertian surfaces, which assumes that the incident light is equally reflected in all directions \cite{lamber1760photometria}.

Given $f$ images and the light source directions, the surface normal vector and albedo values can be estimated for all the observed surface points~\cite{woodham1980photometric}.
Therefore, \eqref{eq:reflectance_function} can be rewritten in matrix form as:
\begin{equation}
	\mat I =  \mat{L} \mat{n} \mat \rho,
\label{eq:ps}
\end{equation}
where matrix $\mat I_{f \times b}$ is the stack of $f$ vectorized images, each containing $b$ pixels.
Matrices $\mat L_{f \times 3}$ and $\mat n_{3 \times b}$ contain the lighting direction and surface normal vectors, respectively.
The diagonal matrix $\mat \rho_{b \times b}$ contains the albedo values for all the pixels.
In case of calibrated PS---with known light source directions---three images are enough to estimate the surface normals.
Because of shadows and not perfect Lambertian properties as described by Sun in \cite{sun2007object}, three images are mostly not enough when working with real data.
Therefore, $f \ge 3$.
Given the light source directions $\mat L$ and the intensity matrix $\mat I$, albedos and surface normals can be calculated as:
\begin{eqnarray}
	\mat \rho = diag(\vert \mat{L}^{-1}\mat{I} \vert) 
	\label{eq:rho},\\
	\mat{n} = \frac{1}{\mat{\rho}}\mat{L}^{-1}\mat{I} 
	\label{eq:normal}.
\end{eqnarray}

\section{Designing the LightPanel}
\label{sec:lightpanel}

Our image acquisition system, the \textit{LightPanel}, consists of a camera, a distance sensor and several light sources---called \textit{LightBlob}---on the circumference of a circle around the camera.
Every \textit{LightBlob} has 2-DOF, which are used to change their distance from the camera and change the emitting angle.
Using such configurable \textit{LightBlob}s allow the \textit{LightPanel} to have different emitting focal length and consequently be able to capture images from objects at different distances from the camera.
Every \textit{LightBlob} consists of several LEDs assembled in a way to provide proper lighting to the scene.

\subsection{LightBlob}
\label{ssec:lightblob}

\begin{figure}[t]
	\centering
	\centerline{\includegraphics[width=7cm, trim = 30mm 10.5cm 30mm 10.5cm, clip]{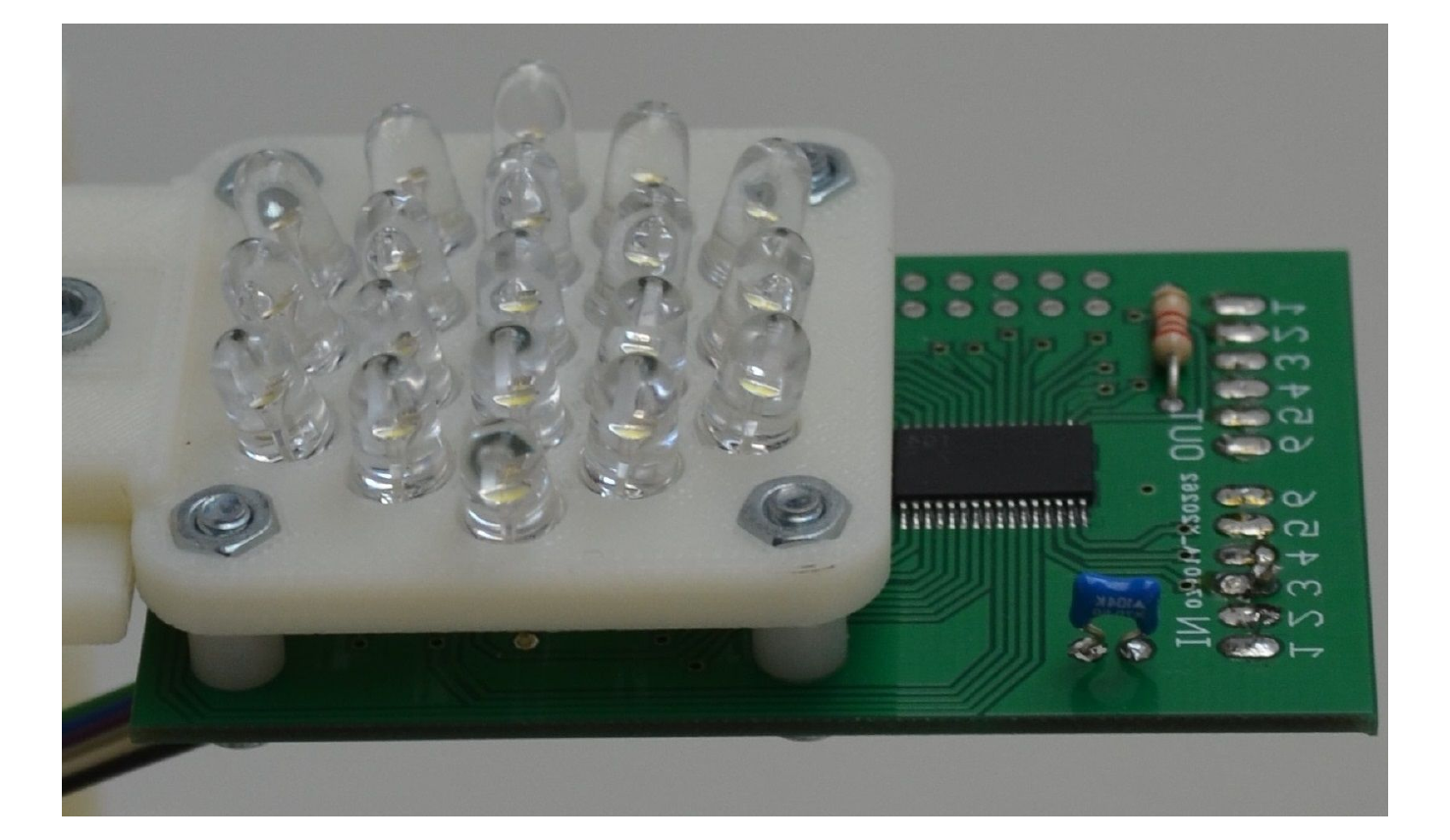}}
	\caption{Arrangement of LEDs in the \textit{LightBlob}.}
	\label{fig:blob_geometry}
\end{figure}

The \textit{LightBlob} is an arrangement of 19 LEDs with viewing angles of $15^\circ$ sitting together to act as single light source that has a wider viewing angle and uniformly illuminates the scene.
Such arrangement satisfies the standard calibrated PS algorithm assumption of parallel light rays from a single distant light source.
The advantage of a small opening angle is that the light direction is defined by the orientation of LEDs, but indeed the covered area is smaller.
Therefore, in every \textit{LightBlob} multiple bright LEDs are arranged in a way that they jointly build a cone and cover a bigger area.
The luminous intensity of a single LED is $35 Cd$ at a current of $20 mA$ \cite{seoul2014led}.
The relative luminous intensity depending on the forward current as well as the radiation properties are considered to find the angles between LEDs to achieve a smooth intensity on the object.
Considering the LEDs specifications the angles between the LEDs is defined to be $11.25^\circ$, as it is shown in the Figure~\ref{fig:blob_geometry}.
The electronic circuit that derives the \textit{LightBlob} allow for vary its viewing angle between $30^\circ$ and $52.5^\circ$, by turning the outer ring of LEDs on and off.
This allows to have parallel light rays even at close distances.

\subsection{LightPanel}
\label{ssec:lightpanel}

\begin{figure}[t]
	\centering
	\centerline{\includegraphics[width=9cm, trim = 0mm 15mm 0mm 15mm, clip]{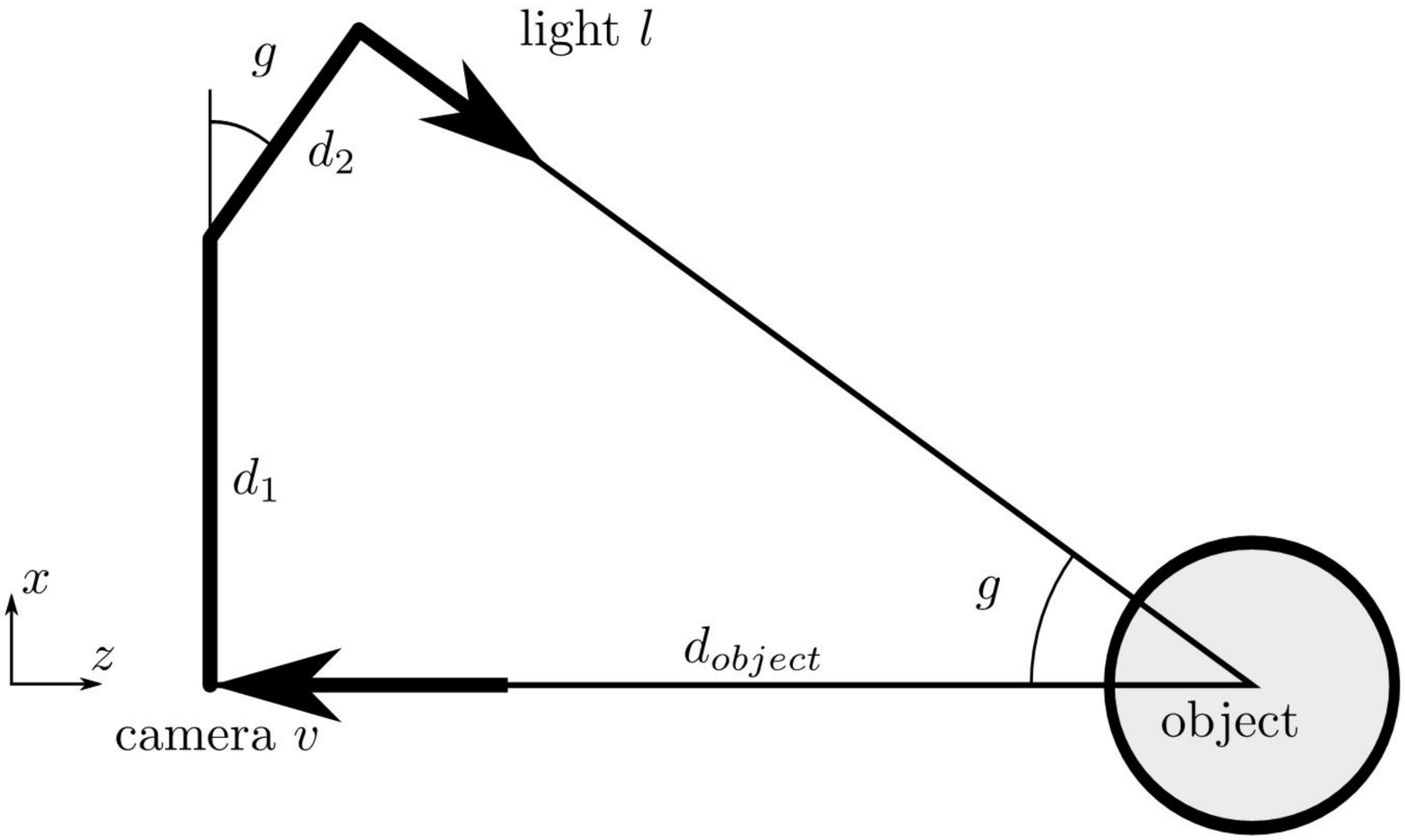}}
	\caption{Illustration of a wing in the \textit{LightPanel}.}
	\label{fig:light_panel_angle}
\end{figure}

The main body of the \textit{LightPanle} is a disc with the camera in the middle.
Around the disc, several wings, each having a \textit{LightBlobs}, are attached.
The design allows adding up to twelve wings.
Every wing consists of spacers and a joint to align the \textit{LightBlob}s with a certain focal point.
Using the joint, the angle of light direction can be varied from $10^\circ$ to $80^\circ$ in steps of $10^\circ$.
Combination of the length of different spacers and the joint angles define the focal point of the light panel.
The distance between camera and the focal point is the operating distance of the light panel and is called $d_{object}$.
The schematic of such a distance is illustrated in Fig.~\ref{fig:light_panel_angle}.

The distance $d_{object}$ is defined by the phase angle $g$ and the distances $d_1$ and $d_2$, where
$d_1 = d_r + d_{spacer} + d_{joint}$ is the sum of the disc radius $d_r$, which holds the camera and the wings, a variable spacer $d_{spacer}$ in the middle and the joint radius $d_{joint}$.
Similarly, $d_2 = d_{joint} + d_{spacer} + d_{blob}$ consists of the joint radius $d_{joint}$, a part of the light blob $d_{blob}$ and a variable spacer $d_{spacer}$.
The distance $d_{object}$ is given by 
\begin{equation}
	d_{object} = \frac{d_1}{\tan(g)}+\frac{d_2}{\sin(g)}.
\label{eq:operatingdistance}
\end{equation}
Therefore, the light source direction $\vec l_s = \left[-\sin(g),\: 0,\: \cos(g)\right]^T$ depends only on the phase angle $g$.

\subsection{Calibration}
\label{ssec:lpcalib}

\begin{figure}
	\centering
	\centerline{\includegraphics[width=\linewidth, trim = 2mm 80mm 2mm 80mm, clip]{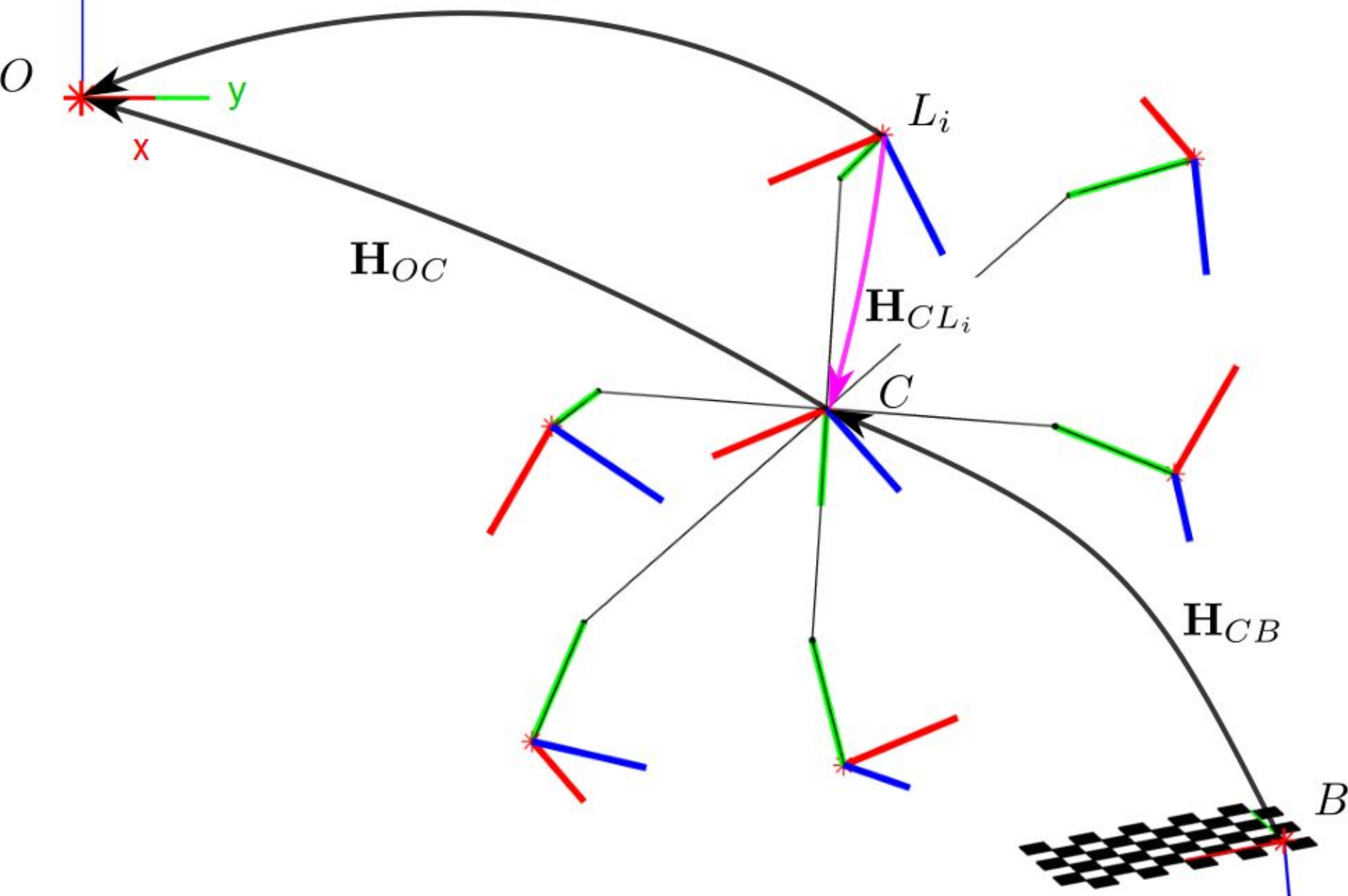}}
	\caption{Transformations of the \textit{LightPanel}. The unknown transformation $\textbf{H}_{CL_i}$ is illustrated in magenta.}
	\label{fig:panel_trafos}
\end{figure}

The aim of calibration process is to estimate the light source directions in the camera frame.
For this reason the camera frame $C$ has to be estimated in the first step.
The light panel $P$ is equipped with markers to track the light panel position in the motion capture system $\mat{H}_{OP}$.
Calibration of the camera provides the checkerboard position in the camera frame $\mat{H}_{CB}$.
Using Matlab calibration toolbox \cite{wengert2014fully}, the transformation $\mat{H}_{PC}$ is obtained. 
Then, the light direction vector for every \textit{LightBlob} $\vec l_i$ ($i = 1 \dots f$), has to be derived.
Such vectors can be obtained by considering the \textit{LightPanel} geometry, so the transformation of light vectors from the light frames to the world reference frame are called $\mat{H}_{OL_i}$.
Therefore, the transformation of the light frames to the camera frame, as illustrated in Fig.~\ref{fig:panel_trafos}, can be defined as:
\begin{equation}
	\mat{H}_{CL_i} = (\mat{H}_{OP}\mat{H}_{PC})^{-1}\mat{H}_{OL_i}.
\end{equation}

\section{Experiments and Discussions}
\label{sec:results}

In this section challenges in designing and using an image acquisition setup are discussed and followed by a few results obtained by using the \textit{LightPanel} in practical applications are presented.

\subsection{How should the light be illuminated?}
\begin{figure}[t]
	\centering
	\centerline{\includegraphics[width=0.8\linewidth, trim = 0mm 5mm 0mm 5mm, clip]{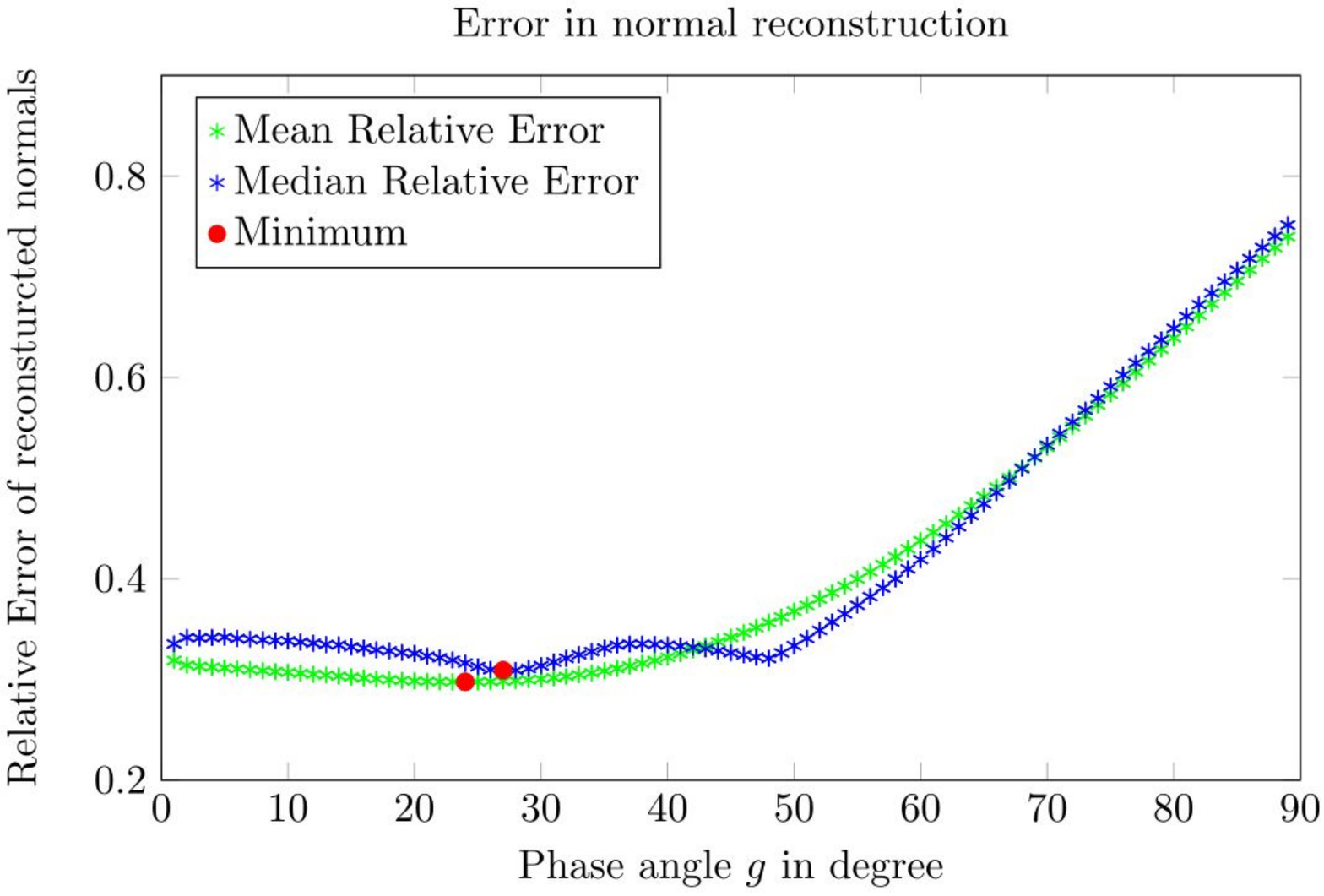}}
\caption{Mean and Median reconstruction error for the surface normals.}
\label{fig:angle_error}
\end{figure}

In order to investigate the effect of light emitting direction on the reconstruction an experiment is designed in a simulating environment.
More specifically, the goal of this experiment is to find what is the best value for phase angle $g$, which is introduced in both Fig.~\ref{fig:ref_geo} and Fig.~\ref{fig:light_panel_angle}.
A synthetic unit sphere with perfect diffuse surface is created and an orthographic camera is placed in front of it.
The light source is chosen to have parallel rays pointing to the centre of the sphere.
The intensity of the light source was chosen in a way that no pixel intensity is saturated.
The light source is rotated $360^\circ$ around the camera and every $60^\circ$ an image was rendered.
The number of images was increased from three images of the standard calibrated photometric stereo algorithm to six images, because with six images of a convex object illuminated by six different light positions it can be ensured that every visible surface point is illuminated by three light sources \cite{sun2007object}.
In this setup the only parameter is the phase angle $g$.
Therefore, the phase angle was varied from $1^\circ$ to $89^\circ$ in steps of $1^\circ$.
Using \eqref{eq:rho} and \eqref{eq:normal} on synthetic data consisting of six images of a diffuse sphere with known illumination and a constant phase angle $g$, 
the estimated normals are compared with the normals of the sphere.
The mean and median relative errors are shown in Fig.~\ref{fig:angle_error}.

\begin{figure}[tb]
	\centering
	\centerline{\includegraphics[width=0.8\linewidth, trim = 0mm 1mm 0mm 1mm, clip]{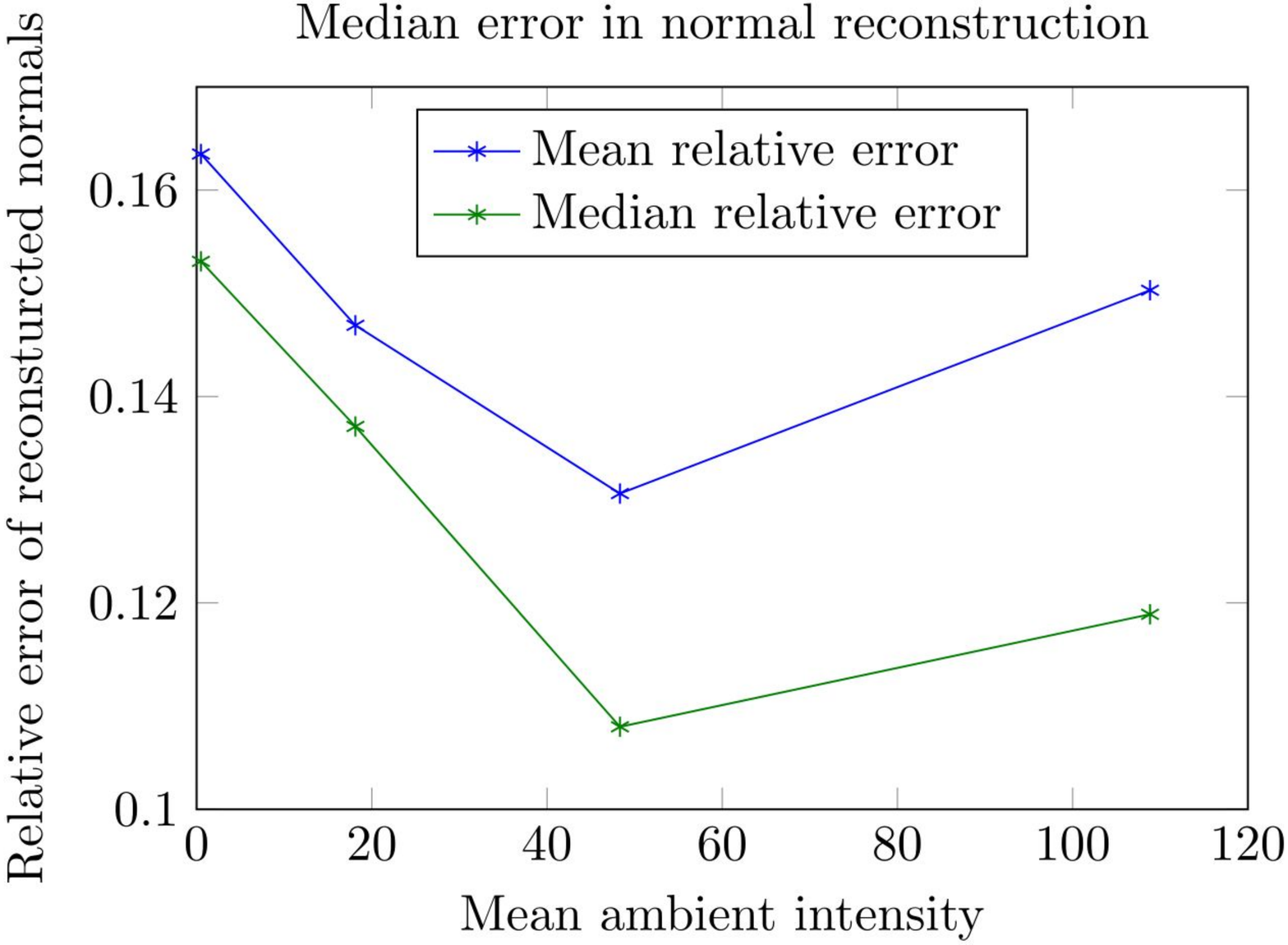}}
	\caption{Mean and median relative error with increasing ambient light.}
	\label{fig:ambienterror}
\end{figure}

\subsection{Is the ambient light disturbing?}
The \textit{LightPanel} should be applicable in the presence of ambient light.
To effectively estimate the surface normals, the ambient light should be removed from the images.
Therefore, as suggested by \cite{jones2011head}, for using the \textit{LightPanel} an image is taken without turning on the \textit{LightBlobs}. 
Then, this image is subtracted from images taken under active lighting.
To investigate the effect of ambient light, a light source which imitates the ambient light is added to the synthetic experiment.
The amount of ambient light is measured with the pixel intensities on a scale from 0 to 255 of the sphere in the ambient light image.
The comparison of the estimations proceeds as in the experiments before.
Fig.~\ref{fig:ambienterror} illustrates the mean and median relative errors against the mean ambient intensity.
The best result of the four datasets is achieved with about $20\%$ ambient light, which shows that the light panel can be used with ambient light.
It means that having some ambient light helps the active light source to illuminate the object smoothly by decreasing the number of shadowed pixels.
However, too much ambient light can disturb the whole estimation if the pixel intensities are saturated with the ambient light.

\begin{figure}[t!]
	\centering
	\centerline{\includegraphics[width=0.8\linewidth, trim = 25mm 0mm 25mm 0mm, clip]{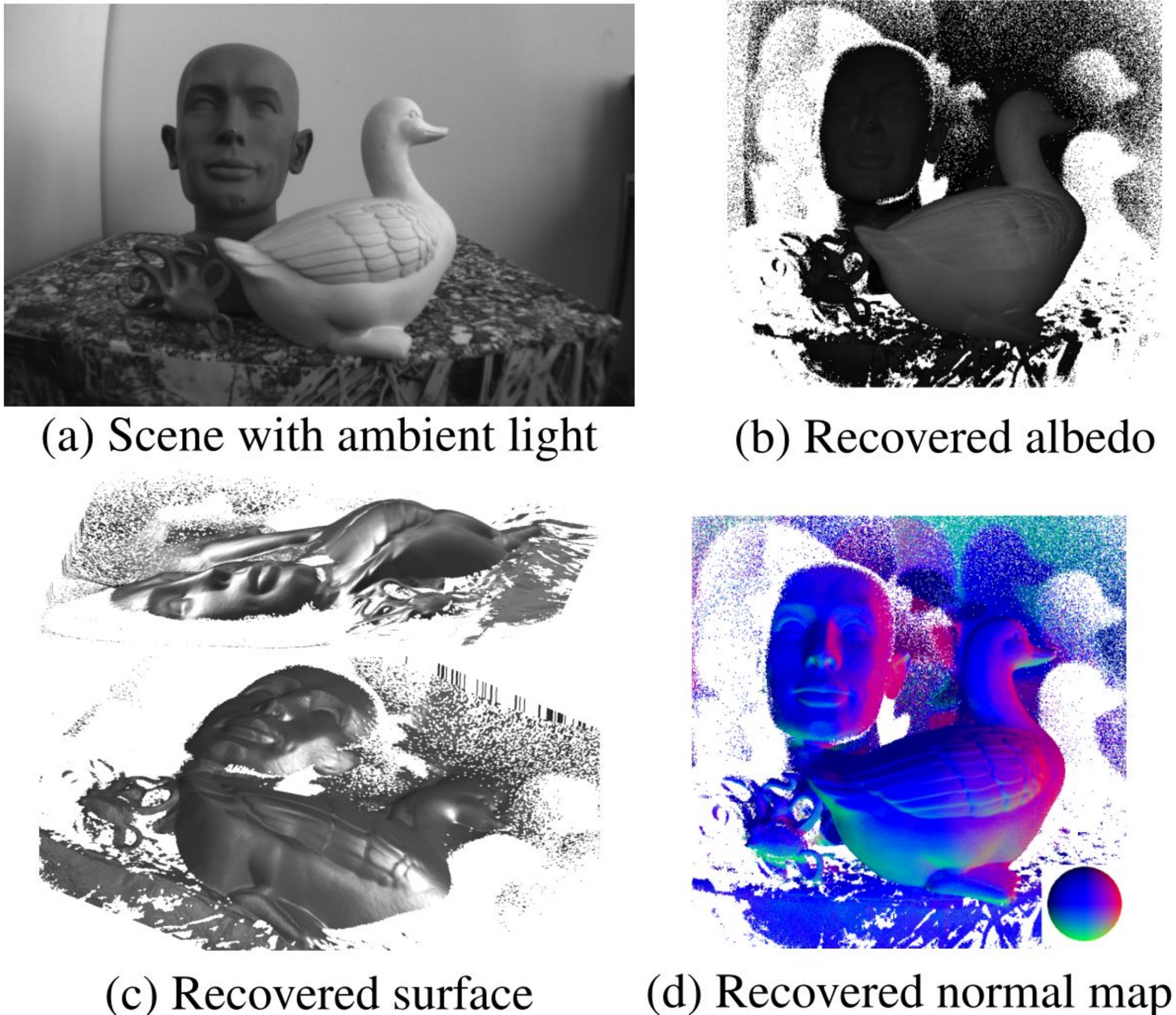}}
	\caption{Reconstruction of scene with multiple objects.}
	\label{fig:results}
\end{figure}

\subsection{Results}
Several experiments have been done using the \textit{LightPanel} to reconstruct objects and scenes~\footnote{More results can be found in \href{https://youtu.be/hUYDQJiKiLU}{this video}.}.
The camera captures $752\times 480$ pixel grayscale images at $12 Hz$.
Fig.~\ref{fig:results} shows one of the most challenging scenarios used for reconstruction.
For this experiment, the phase angle $g = 30^\circ$ and distance $d_{object} = 0.35 m$ and images are captured with $6$ different lighting conditions.
In this experiment, the objects are not segmented from the background and the resulted normal map contains many outliers for the surface points belonging to the background.
Such artefacts in the normal map make the normal integration step---to get surface from normal map---more challenging.
Consequently, the reconstruction of the normal map in Matlab took $14.58 \; Sec$ for $153756 \; Pixels$.
A further problem is the occlusion of the objects in the background, because the occluded points are not sufficiently illuminated.
A possible solution to improve this reconstruction could be to equip the panel with more light sources and try different algorithms for the segmentation of the objects before applying photometric stereo.

\section{Conclusions}
\label{sec:conclusions}
This paper introduces a mobile setup which uses active light and employs PS for dense 3D modelling.
Operating this system in uncontrolled lighting conditions introduces challenges for using PS in practice.
Such challenges are addressed in this paper and supported by experiments on synthetic and real data.
In a nutshell, we confirm that the ambient light can help PS reconstruction by reducing the number of shadowed pixel and providing a smooth normal map. 
Moreover, the incident angle of light rays plays an important role in the quality of reconstruction.
Our experiments shows that the best results can be achieved by setting this angle between $20^\circ$ and $30^\circ$.

\section*{Acknowledgement}
This research was supported by the Hasler Foundation---project number 13027---and the UZH Forschungskredit---project number FK-14-015.

\bibliographystyle{ieeetr} 
\bibliography{refs}

\end{document}